\documentclass[pdflatex,sn-mathphys-num]{sn-jnl}

\usepackage{graphicx}%
\usepackage{multirow}%
\usepackage{amsmath,amssymb,amsfonts}%
\usepackage{amsthm}%
\usepackage{mathrsfs}%
\usepackage[title]{appendix}%
\usepackage{xcolor}%
\usepackage{textcomp}%
\usepackage{manyfoot}%
\usepackage{booktabs}%
\usepackage{algorithm}%
\usepackage{algorithmicx}%
\usepackage{algpseudocode}%
\usepackage{listings}%

\theoremstyle{thmstyleone}%
\theoremstyle{thmstyletwo}%

\theoremstyle{thmstylethree}%

\begin{document}

\title[Article Title]{Vision-Language Models for Automated Chest X-ray Interpretation: Leveraging ViT and GPT-2}


\author*[1]{\fnm{Md. Rakibul} \sur{Islam}}\email{rakib.aust41@gmail.com}
\equalcont{These authors contributed equally to this work.}

\author[1]{\fnm{Md. Zahid} \sur{Hossain}}\email{zahidd16@gmail.com}
\equalcont{These authors contributed equally to this work.}

\author[1]{\fnm{Mustofa} \sur{Ahmed}}\email{mustofahmed24@gmail.com}
\equalcont{These authors contributed equally to this work.}

\author[2]{\fnm{Most. Sharmin Sultana} \sur{Samu}}\email{sharminsamu130@gmail.com}
\equalcont{These authors contributed equally to this work.}

\affil[1]{\orgdiv{Department of Computer Science and Engineering}, \orgname{Ahsanullah University of Science and Technology}, \orgaddress{\city{Dhaka}, \postcode{1208}, \country{Bangladesh}}}

\affil[2]{\orgdiv{Department of Civil Engineering}, \orgname{Ahsanullah University of Science and Technology}, \orgaddress{\city{Dhaka}, \postcode{1208}, \country{Bangladesh}}}


\abstract{Radiology plays a pivotal role in modern medicine due to its non-invasive diagnostic capabilities. However, the manual generation of unstructured medical reports is time consuming and prone to errors. It creates a significant bottleneck in clinical workflows. Despite advancements in AI-generated radiology reports, challenges remain in achieving detailed and accurate report generation. In this study we have evaluated different combinations of multimodal models that integrate Computer Vision and Natural Language Processing to generate comprehensive radiology reports. We employed a pretrained Vision Transformer (ViT-B16) and a SWIN Transformer as the image encoders. The BART and GPT-2 models serve as the textual decoders. We used Chest X-ray images and reports from the IU-Xray dataset to evaluate the usability of the SWIN Transformer-BART, SWIN Transformer-GPT-2, ViT-B16-BART and ViT-B16-GPT-2 models for report generation. We aimed at finding the best combination among the models. The SWIN-BART model performs as the best-performing model among the four models achieving remarkable results in almost all the evaluation metrics like ROUGE, BLEU and BERTScore.}




\maketitle

\section{Introduction}\label{sec1}

Medical reports play a key role in summarizing the diagnoses observed in X-ray images. These reports are essential for promoting efficient communication among healthcare professionals. They help guarantee precise care and facilitate well-informed decision-making. However, creating medical reports requires considerable expertise and time as radiologists typically write them manually. It can take a lot of time and effort to complete this manual process. Automating medical report generation offers the potential to improve patient care, streamline administrative workflows and enhance overall healthcare delivery. Consequently, several studies \cite{chen2022cross} \cite{jing2017automatic} \cite{li2019knowledge} \cite{tanida2023interactive} have focused on developing automated models to generate medical reports with most current efforts dedicated to chest radiology reports.

Despite advancements, significant challenges hinder progress in this field. Accurately interpreting medical images remains a complex task compounded by the need for meticulous data annotation and addressing heterogeneity in input data. Ensuring consistency and standardization in generated reports is also a major hurdle. Furthermore, the diversity and variability of diseases along with the requirement for algorithm interpretability present additional obstacles. Overcoming these challenges is crucial for improving the quality and reliability of automated medical report generation and makes this an urgent area of research.

Automated radiology report generation systems are designed to process medical images and create textual reports that describe radiological findings and interpretations. These systems aim to transform radiology practice by automating repetitive tasks, identifying potential medical conditions and reducing diagnostic errors. Such systems have the potential to enhance workflows by prioritizing patients based on the urgency of their conditions in clinical settings. This capability could significantly improve operational efficiency and save lives by expediting critical diagnoses.

Recent research has demonstrated that transformer-based models and global visual features deliver promising results in automated report generation. However, these models often struggle with identifying and characterizing rare or complex diseases. The primary limitation lies in their inadequate understanding of medical terminology, anatomical structures and lesion characteristics. As a result, the generated descriptions frequently lack professionalism and lead to text that may be unclear or insufficiently fluent. These issues compromise the usability and practical application of the reports in clinical settings.

To address these challenges we use deep learning models to combine textual and visual input. We employ the Vision Transformer (ViT-B16) and SWIN Transformer as encoders, BART and GPT-2 as decoders. This architecture combines the strengths of advanced vision and language models with the aim to enhance the overall quality of automated medical report generation. 

We conducted a systematic evaluation of different encoder-decoder combinations including ViT B16 with BART, SWIN Transformer with BART, SWIN Transformer with GPT-2 and ViT B16 with GPT-2. Each configuration was assessed based on its ability to improve disease recognition accuracy, enhance the professionalism of disease descriptions and ensure fluency in report text. Our goal is to maximize the automation of radiology report generation by determining the most efficient combination.
The following concisely describes our main contributions:
\begin{itemize}
\item  We introduce an approach that integrates visual and textual data using Vision Transformers (ViT B16) and SWIN Transformers as encoders along with BART and GPT-2 as decoders.
\item We systematically evaluate four encoder-decoder combinations (SWIN Transformer-BART, SWIN Transformer-GPT-2, ViT B16-BART, ViT B16-GPT-2) to identify the optimal pairing for addressing key challenges such as disease recognition accuracy, professionalism in descriptions and text fluency.
\item Our method focuses on reducing the burden on clinicians by automating the report-writing process by improving diagnostic precision and ensuring that generated reports adhere to professional medical standards.
\end{itemize} 

The structure of this article is as follows. Section 2 presents a summary of related studies. Section 3 contains the methodology of our approach. Section 4 describes the experimental setup along with the description of the dataset and the pre-processing of the data. It also reveals the hyperparameter combinations for different models. Section 5 reports the analysis of the research results with a comparison among the four multimodal models. Lastly, Section 6 concludes the article with discussion about the limitations and future work directions.

\section{Related Work}\label{sec2}

The studies explore different architectures for automated report generation. \cite{lovelace2020learning} introduces a transformer-based model with a differentiable clinical information extraction approach. In contrast, \cite{hou2021automatic} employs an adversarial reinforcement learning framework integrating an accuracy discriminator (AD) and a fluency discriminator (FD). \cite{sirshar2022attention} combines Convolutional Neural Networks (CNNs) and Long Short-Term Memory (LSTM) networks with an attention mechanism to improve feature extraction and text generation. \cite{ahmed2022explainable} focuses on explainable AI (XAI) techniques utilizing class activation maps (CAM) for image analysis and LIME for explainability in natural language processing. \cite{li2022self} proposes a self-guided framework (SGF) that combines unsupervised and supervised deep learning techniques allowing label-free learning from medical reports.
The datasets used in these studies include MIMIC-CXR \cite{johnson2019mimic}, Indiana University (IU) X-ray \cite{demner2016preparing} and other medical report repositories. \cite{lovelace2020learning} and \cite{hou2021automatic} utilize both MIMIC-CXR \cite{johnson2019mimic} and IU X-ray \cite{demner2016preparing} datasets to ensure robustness in evaluation. \cite{sirshar2022attention} also uses these datasets emphasizing feature extraction through CNNs and LSTMs. \cite{ahmed2022explainable} compiles data from various studies rather than using a specific dataset. \cite{li2022self} primarily relies on existing radiology reports without requiring annotated disease labels that take advantage of text image associations to refine its self-guided learning process.
\cite{lovelace2020learning} reports improvements in clinical coherence achieving a 6.4-point increase in micro-averaged F1 scores and a 2.2-point gain in macro-averaged F1 scores. \cite{hou2021automatic} demonstrates superior natural language generation (NLG) performance on the IU X-ray \cite{demner2016preparing} and MIMIC-CXR \cite{johnson2019mimic} datasets through adversarial reinforcement learning. \cite{sirshar2022attention} achieves state-of-the-art BLEU scores outperforming LRCN and hierarchical RNNs by incorporating attention mechanisms. \cite{li2022self} shows significant improvements in report accuracy and length by demonstrating a better alignment between image features and textual descriptions. However, \cite{ahmed2022explainable} does not present empirical results but highlights gaps in the application of XAI techniques to medical NLP.
\cite{lovelace2020learning} acknowledges that clinical coherence remains inadequate for real-world deployment. \cite{hou2021automatic} notes that the small size of the IU X-ray dataset and the structural simplicity of reports may affect model performance. \cite{sirshar2022attention} struggles with complex disease presentations and coherent long-form text generation. \cite{ahmed2022explainable} identifies the lack of a universal XAI method applicable to both vision and NLP models as a major gap. \cite{li2022self} points out challenges in handling medical language complexity and potential biases in text-image associations.
\cite{lovelace2020learning} suggests integrating retrieval-based methods to improve clinical accuracy. \cite{hou2021automatic} proposes expanding dataset diversity and incorporating additional evaluation metrics. \cite{sirshar2022attention} recommends optimizing attention mechanisms and enhancing NLP capabilities for better long-form text generation. \cite{ahmed2022explainable} highlights the need for hybrid XAI techniques and interdisciplinary collaboration. \cite{li2022self} aims to refine its self-guided learning approach by integrating more diverse data sources and improving the model’s understanding of complex medical language.

\cite{wang2022medical} introduces a Medical Semantic-Assisted Transformer that incorporates a memory-augmented sparse attention block and a Medical Concepts Generation Network (MCGN) to enhance semantic coherence. \cite{li2023harnessing} presents MedEPT, a parameter-efficient approach leveraging pre-trained vision-language models through a Generator and Discriminator framework with knowledge extraction using MetaMap. \cite{nicolson2023improving} employs a multi-modal machine learning approach that integrates transfer learning from pre-trained models to combine image and text representations. \cite{voutharoja2023automatic} focuses on contrastive learning by synthesizing increasingly hard negatives to improve feature discrimination without adding extra network complexity. \cite{tanida2023interactive} proposes a region-guided model that detects and describes specific anatomical regions enabling improved accuracy and human interaction.
\cite{wang2022medical} employs the MIMIC-CXR \cite{johnson2019mimic} dataset, \cite{li2023harnessing} uses image-only datasets reducing reliance on extensive annotated data, \cite{nicolson2023improving} utilizes chest X-ray report generation datasets, \cite{voutharoja2023automatic} employs both the IU-XRay \cite{demner2016preparing} and MIMIC-CXR \cite{johnson2019mimic} datasets to improve model generalization and \cite{tanida2023interactive} demonstrates its effectiveness through experimental evaluations.
\cite{wang2022medical} reports superior performance in image captioning and report generation by integrating medical semantics. \cite{li2023harnessing} significantly reduces trainable parameters and training time while maintaining high accuracy making medical report generation more efficient. \cite{nicolson2023improving} effectively combines image and text representations and improves chest X-ray caption accuracy. \cite{voutharoja2023automatic} enhances feature discrimination using hard negatives and achieves state-of-the-art alignment between images and reports. \cite{tanida2023interactive} improves report quality and transparency of the report by incorporating anatomical region guidance enabling better clinical applicability.
\cite{wang2022medical} highlights the complexity of radiographic images as a limitation suggesting refinement of model and exploring additional datasets for better generalization. \cite{li2023harnessing} notes its reliance on existing datasets and the potential biases present in medical reports recommending expansion of medical datasets and improving entity extraction methods. \cite{nicolson2023improving} suggests further improvements in pre-trained model applications. \cite{voutharoja2023automatic} acknowledges that generated reports are not yet at human-level quality and proposes enhancing model architecture and training strategies. \cite{tanida2023interactive} suggests improving interactive features and applying the model to various imaging modalities to increase its clinical usability.

\cite{nimalsiri2023automated} introduces MERGIS, a transformer-based model that integrates image segmentation to refine feature extraction significantly improving the coherence of the report and outperforming previous methods in the MIMIC-CXR dataset. However, it requires further validation on diverse datasets. \cite{kumar2023deep} employs a deep learning-based encoder-decoder model with an attention mechanism leveraging CheXnet for feature extraction and demonstrates promising BLEU score results. However, its reliance on limited datasets restricts its generalization potential. \cite{deria2024inverge} presents InVERGe, a lightweight transformer using the Cross-Modal Query Fusion Layer (CMQFL) to bridge visual and textual modalities. Trained on MIMIC-CXR, IU-XRay and CDD-CESM \cite{khaled2022categorized} datasets, it enhances image-text alignment, yet dataset diversity remains a concern. \cite{hoque2024medical} explores large multimodal models integrating LLaVA, IDEFICS 9B and visionGPT2. It achieves strong BERTScore and ROUGE results on the ROCOv2 dataset. However, challenges in concept detection persist suggesting the need for additional datasets and refined models. \cite{magalhaes2024xrayswingen} introduces XRaySwinGen, a multimodal model combining the SWIN Transformer for image encoding and GPT-2 for text generation with bilingual capabilities. Despite high ROUGE-L and METEOR scores, dataset bias and validation across different imaging modalities remain challenges.

\cite{dawidowicz2024image} introduces VLScore, a novel multi-modal evaluation metric that jointly embeds visual and textual features to assess the diagnostic relevance of generated reports. This approach addresses the limitations of prior evaluation methods that rely solely on text-based metrics. Using the ReXVal \cite{yu2023radiology} dataset and a custom perturbation dataset, VLScore demonstrates strong correlation with radiologist evaluations. However, its reliance on dataset-specific constants limits generalizability. \cite{wang2024cxpmrg} proposes CXPMRG-Bench, a benchmarking framework for X-ray report generation using the CheXpert Plus \cite{chambon2024chexpert} dataset. It incorporates a multi-stage pre-training strategy with self-supervised contrastive learning and fine-tuning to enhance report accuracy. Despite its improvements, the framework still requires dataset expansion and architectural refinements for broader applicability. \cite{singh2024designing} presents a robust radiology report generation system integrating multimodal learning combining CNNs for image analysis with RNNs or Transformers for text generation. The model is trained on IU-CXR and leverages pre-trained models and transfer learning to produce clinically coherent reports. However, challenges persist in ensuring full clinical accuracy and real-world validation. \cite{cheddi2024multi} introduces a multi-modal feature fusion-based approach for chest X-ray report generation. By integrating a vision transformer for visual feature extraction with Word2Vec for semantic textual features, this model achieves state-of-the-art performance on the IU-Xray \cite{demner2016preparing} and NIH \cite{wang2017chestx} datasets. Despite its success, it requires further dataset diversification and adaptation for various diseases.\\ \\
The following research gaps are identified through our extensive literature search:
\begin{itemize}
\item  Need for diverse datasets and improvements in architecture.
\item Achieving full clinical accuracy and validating the system in real settings.
\item Need for better feature extraction and validation on diverse datasets.
\item Generated reports are not yet at human-level performance and require further refinement.
\end{itemize}

\section{Methodology}\label{sec3}

\begin{figure}[h]
  \centering
  \includegraphics[width=0.9\textwidth]{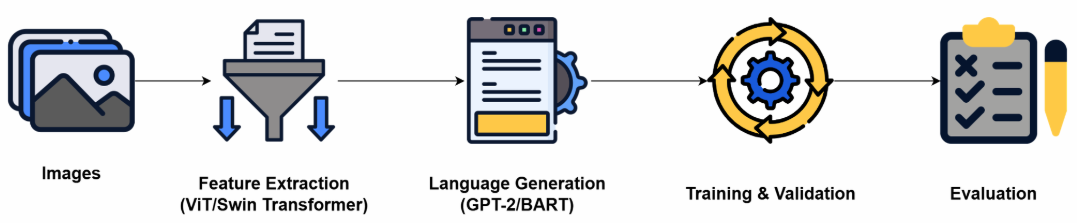}
  \caption{Proposed methodology for automated X-ray interpretation}\label{fig1}
\end{figure}

Our workflow (Fig.\ref{fig1}) begins with image acquisition and processing. Chest X-ray images were collected from publicly available IU-Xray dataset to ensure diversity and relevance for the task. Feature extraction was performed using ViT (Vision Transformer) and SWIN Transformer. The Vision Transformer \cite{mao2022towards} divides the image into fixed-size patches, embeds them into feature vectors and processes these vectors through a series of self-attention layers to capture global contextual information. In contrast, the SWIN Transformer \cite{liu2021swin} employs a hierarchical structure with shifted windows by enabling efficient computation of both local and global image representations. Both models were pre-trained on large-scale image datasets such as ImageNet and subsequently fine-tuned on the medical domain data to extract high-quality features relevant for the task. These extracted features serve as the visual input to the subsequent language generation model.

The language generation stage employs either GPT-2 or BART as the core text generation model. GPT-2 \cite{radford2019language} is a decoder-only architecture that generates descriptive text in an autoregressive manner based on the extracted visual features ensuring coherence and relevance to the input image. BART \cite{lewis2019bart} is a sequence-to-sequence model that combines bidirectional encoding and autoregressive decoding to generate well-structured medical reports. Cross-attention mechanisms were employed to integrate the visual features into the language model to ensure that the generated text is grounded in the visual input. The models were trained using paired image-report datasets to enable them to learn mappings between visual features and corresponding textual descriptions effectively.\\
The training process was conducted in an end-to-end manner using supervised learning. Cross-entropy loss was employed to optimize the model. Validation datasets were utilized during training to monitor model performance and prevent overfitting. The optimization process leveraged the AdamW optimizer with weight decay to ensure convergence.\\
Evaluation of the generated reports was carried out using a mix of textual and image-aware metrics. These include ROUGE \cite{lin2004rouge}, BLEU \cite{papineni2002bleu}, BERTScore \cite{zhang2019bertscore}. More specifically, BLEU assesses text quality through n-gram matching. ROUGE-L evaluates text using the longest common subsequence. BERTScore Precision, BERTScore Recall and BERTScore F1 measure text similarity using contextual embeddings.

\section{Experimental Setup}\label{sec4}

The experiments were carried out using an NVIDIA T4 GPU made available through Google Colab. This GPU provided the necessary computational power to ensure the efficient training and evaluation of the proposed models. The computational setup for the study included standard deep learning libraries. PyTorch was utilized for implementing the models, while Hugging Face Transformers were employed for fine-tuning.

\subsection{Dataset}\label{subsec1}
\begin{figure}[h]
  \centering
  \includegraphics[width=0.85\textwidth]{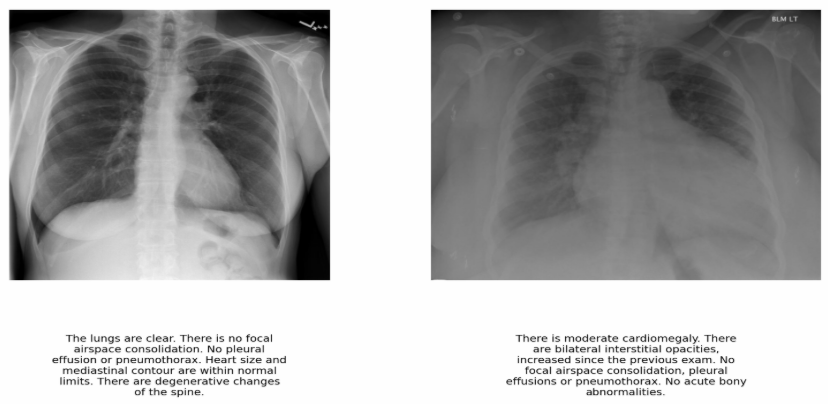}
  \caption{Sample X-ray images and corresponding findings in the form of reports from the IU-Xray dataset. These reports are treated as the ground truth.  }\label{fig2}
\end{figure}

We have used the IU-Xray \cite{demner2016preparing} dataset which is a publicly available collection of radiographic images paired with their corresponding radiology reports. This dataset is widely used for research in medical imaging. The dataset comprises a total of 5,910 chest X-ray images along with their associated findings in the form of radiology reports. Each image in the dataset is accompanied by a detailed textual description that provides diagnostic insights. Fig. \ref{fig2} shows two sample X-ray images with associated reports.

\begin{figure}[h]
  \centering
  \includegraphics[width=0.68\textwidth]{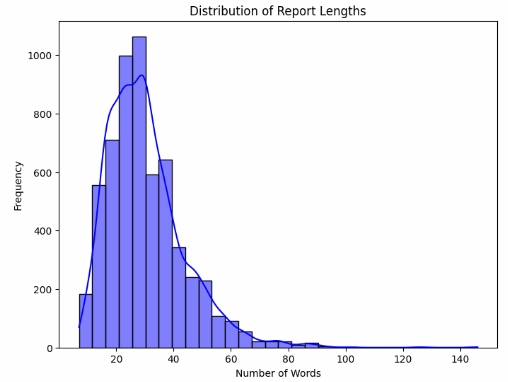}
  \caption{Distribution of report length in number of words}\label{fig3}
\end{figure}

\begin{figure}[h]
  \centering
  \includegraphics[width=0.68\textwidth]{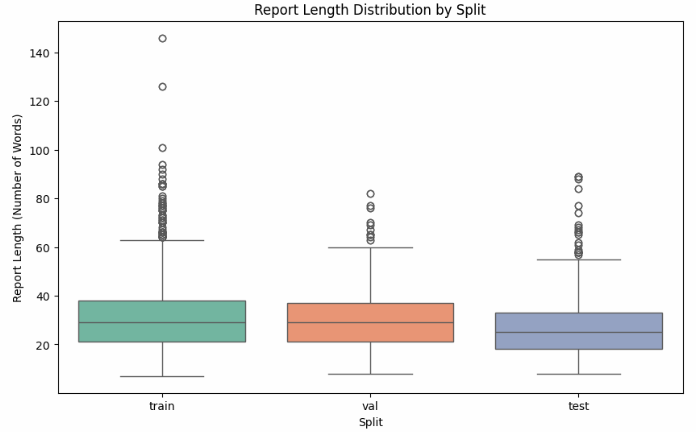}
  \caption{Report length distribution in train, test and validation split}\label{fig4}
\end{figure}

The dataset is organized into predefined splits for training, testing and validation. Training set, test set and validation set contain 4138, 1180 and 592 images and their corresponding reports respectively. The predefined split ratio adheres to a 70:20:10 distribution for training, testing and validation respectively. The dataset is free from missing images, reports or split information. Figures \ref{fig3} and \ref{fig4} show the distribution of report length by the number of words and the distribution of report length in the train-test-validation split respectively.

\subsection{Data Preprocessing and Analysis}\label{subsec2}
\begin{figure}[h]
  \centering
  \includegraphics[width=0.65\textwidth]{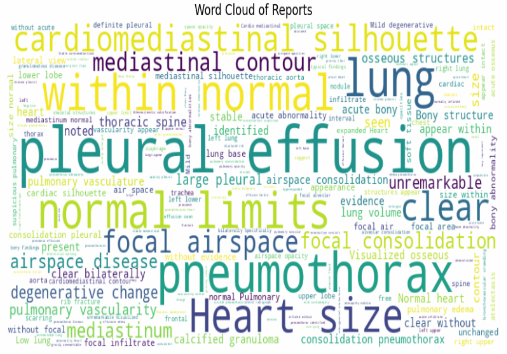}
  \caption{Word cloud of reports before removing stopwords}\label{fig5}
\end{figure}

\begin{figure}[h]
  \centering
  \includegraphics[width=0.65\textwidth]{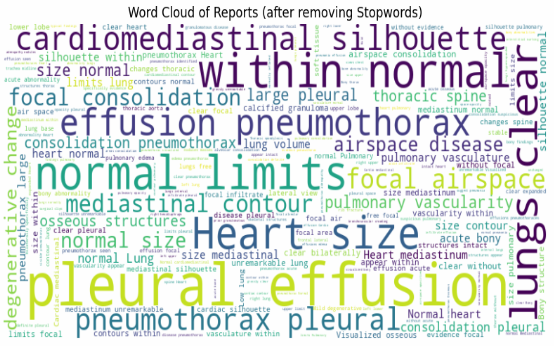}
  \caption{Word cloud of reports after removing stopwords}\label{fig6}
\end{figure}

As part of the preprocessing pipeline, we have removed stopwords systematically from the radiology reports in the dataset. Fig. \ref{fig5} and Fig. \ref{fig6} show the word cloud of reports before and after removing stopwords respectively.

Table. \ref{Report Length by Split} lists detailed statistics of report length in train, test and validation split. The training set contains 4,138 samples, the test set contains 1,180 samples and the validation set contains 592 samples. The mean value indicates a balanced distribution across all three subsets.

\begin{table}[h]
  \caption{Statistics of Report Length by Split}\label{Report Length by Split}
  \begin{tabular}{@{}lllcccccc@{}}
   \toprule
\textbf{Split} & \textbf{Count} & \textbf{Mean} & \multicolumn{1}{c}{\textbf{Standard Deviation}} & \textbf{Min} & \textbf{Max} & \textbf{25\%} & \textbf{50\%} & \textbf{75\%} \\
    \midrule
    Train & 4138.0 & 31.765 & 14.206 & 7.0 & 149.0 & 22.0 & 29.0 & 39.0\\
    
    Test & 1180.0 & 28.219 & 13.181 & 8.0 & 93.0 & 19.0 & 25.0 & 33.0\\
    
    Validation & 592.0 & 31.128 & 13.812 & 8.0 & 83.0 & 21.0 & 30.0 & 38.0\\
    \botrule
\end{tabular}
\end{table}

\subsection{Model Hyperparameters}\label{subsec3}

\begin{table}[h]
  \caption{Model Hyperparameters}\label{Model Hyperparameters}
  \begin{tabular}{@{}lccccc@{}}
   \toprule
\textbf{Model} & \multicolumn{1}{c}{\textbf{Number of}} & \textbf{Training Batch} & \textbf{Optimizer} & \textbf{Learning} & \textbf{Weight} \\
               & \multicolumn{1}{c}{\textbf{Epochs}}     & \textbf{Size} &                  & \textbf{Rate}     & \textbf{Decay}  \\
    \midrule
    ViT B16-GPT-2 & 5 & 8 & AdamW & 0.00005 & 0.01 \\

    ViT B16-BART & 8 & 8 & AdamW & 0.00005 & 0.01 \\
    
    SWIN-BART & 8 & 8 & AdamW & 0.00005 & 0.01 \\
    
    SWIN-GPT-2 & 5 & 8 & AdamW & 0.00005 & 0.01 \\
    \botrule
\end{tabular}
\end{table}

Table \ref{Model Hyperparameters} summarizes the hyperparameters used for training different model architectures in our experiment. Each model is trained using the AdamW optimizer with a consistent learning rate of 0.00005 and a weight decay of 0.01. The training batch size is fixed at 8 across all models. The number of training epochs varies. ViT B16-GPT-2 and SWIN-GPT-2 are trained for 5 epochs, while ViT B16-BART and SWIN-BART are trained for 8 epochs.

\section{Result Analysis}\label{sec5} 
We evaluated the performance of four vision-language models such as SWIN Transformer-BART, SWIN Transformer-GPT-2, ViT B16-BART and ViT B16-GPT-2 on automated X-ray interpretation tasks. The models were compared across multiple metrics including ROUGE scores, BLEU score and BERTScore to assess their syntactic precision, contextual fidelity and semantic similarity.

Table. \ref{Model Evaluation-1} shows that, the SWIN-BART model demonstrated superior performance across all ROUGE metrics. This indicates the effectiveness in capturing textual overlap and contextual accuracy. For ROUGE1 F1, SWIN-BART achieved the highest score of 0.4134 significantly outperforming ViT B16-GPT-2 with a score of 0.2877, ViT B16-BART with a score of 0.3176 and SWIN-GPT-2 with a score of 0.2855.

\begin{table}[h]
  \caption{Model Evaluation with ROUGE and BLEU score}\label{Model Evaluation-1}
  \begin{tabular}{@{}lcccccc@{}}
   \toprule
   \textbf{Model} & \multicolumn{1}{c}{\textbf{ROUGE1}} & \textbf{ROUGE2} & \textbf{ROUGE3} & \textbf{ROUGE4} & \textbf{ROUGEL} & \textbf{BLEU} \\
               & \multicolumn{1}{c}{\textbf{F1}}     & \textbf{F1} &  \textbf{F1}                 & \textbf{F1}     & \textbf{F1}  \\
    \midrule
    ViT B16-GPT-2 & 0.2877 & 0.1273 & 0.0689 & 0.0435 & 0.2031 & 0.0403\\

    ViT B16-BART & 0.3176 & 0.0612 & 0.0121 & 0.0029 & 0.2324 & 0.0169\\
    
    SWIN-BART & 0.4134 & 0.1537 & 0.0738 & 0.0427 & 0.2935 & 0.0648\\
    
    SWIN-GPT-2 & 0.2855 & 0.1108 & 0.0531 & 0.0305 & 0.1933 & 0.0319\\
    \botrule
\end{tabular}
\end{table}

In bigram and trigram overlaps measured by ROUGE2 F1 and ROUGE3 F1, SWIN-BART scored 0.1537 and 0.0738 respectively. These values were notably higher compared to ViT B16-GPT-2, which scored 0.1273 in ROUGE2 F1 and 0.0689 in ROUGE3 F1. Similarly, ViT B16-BART achieved 0.0612 in ROUGE2 F1 and 0.0121 in ROUGE3 F1, while SWIN-GPT-2 had 0.1108 in ROUGE2 F1 and 0.0531 in ROUGE3 F1.\\
For ROUGE4 F1, which evaluates four-gram overlaps, SWIN-BART maintained its lead with a score of 0.0427, while ViT B16-GPT-2 scored 0.0435, ViT B16-BART scored 0.0029 and SWIN-GPT-2 scored 0.0305. In ROUGEL F1, SWIN-BART achieved the highest score of 0.2935, surpassing ViT B16-GPT-2 at 0.2031, ViT B16-BART at 0.2324 and SWIN-GPT-2 at 0.1933. These results establish SWIN-BART as the most capable model in preserving syntactic structure and capturing long-sequence dependencies.\\
The BLEU score, which measures n-gram precision, further confirmed SWIN-BART's dominance. It achieved a score of 0.0648, significantly outperforming ViT B16-GPT-2 at 0.0403, ViT B16-BART at 0.0169 and SWIN-GPT-2 at 0.0319. This result highlights SWIN-BART's enhanced ability to generate text sequences that closely align with reference descriptions.

\begin{table}[h]
  \caption{Model Evaluation with BERTScore}\label{Model Evaluation-2}
  \begin{tabular}{@{}lccc@{}}
   \toprule
\textbf{Model} & \textbf{BERTScore Precision} & \textbf{BERTScore Recall} & \textbf{BERTScore F1} \\
    \midrule
    ViT B16-GPT-2 & 0.8392 & 0.9015 & 0.8691\\

    ViT B16-BART & 0.8158 & 0.8471 & 0.831\\
    
    SWIN-BART & 0.8855 & 0.8947 & 0.8899\\
    
    SWIN-GPT-2 & 0.8331 & 0.8998 & 0.865\\
    \botrule
\end{tabular}
\end{table}

Table. \ref{Model Evaluation-2} shows that, the SWIN-BART model demonstrated superior performance across all BERTScore metrics. This indicates the effectiveness in capturing semantic similarity and contextual relevance in X-ray report generation. For BERTScore Precision, SWIN-BART achieved the highest score of 0.8855, outperforming ViT B16-GPT-2 with a score of 0.8392, ViT B16-BART with 0.8158 and SWIN-GPT-2 with 0.8331. This suggests that SWIN-BART generates text with higher lexical accuracy and better alignment with reference reports.\\

\begin{figure}[h]
  \centering
  \includegraphics[width=0.75\textwidth]{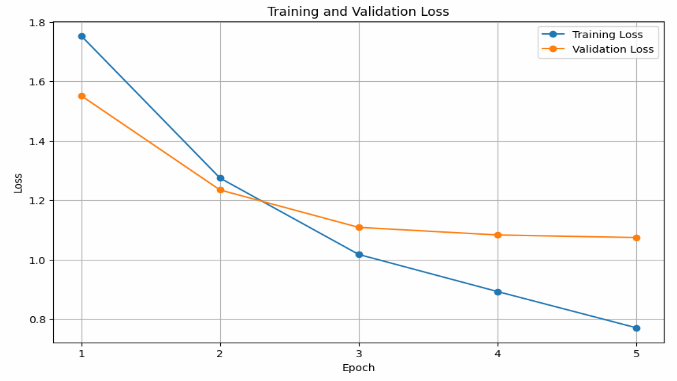}
  \caption{Training vs Validation Loss of ViT B16-GPT-2 Model}\label{fig7}
\end{figure}

In BERTScore Recall, which measures the ability to retain important details from reference reports, SWIN-BART maintained a competitive score of 0.8947, slightly surpassing ViT B16-GPT-2 at 0.9015 and SWIN-GPT-2 at 0.8998. However, it significantly outperformed ViT B16-BART, which scored 0.8471. This highlights the advantage in capturing more clinically relevant details.\\
For BERTScore F1, which balances precision and recall, SWIN-BART achieved the highest score of 0.8899, surpassing ViT B16-GPT-2 at 0.8691, ViT B16-BART at 0.8310 and SWIN-GPT-2 at 0.8650. This confirms its overall effectiveness in generating fluent, coherent and semantically accurate medical reports.

\begin{figure}[h]
  \centering
  \includegraphics[width=0.75\textwidth]{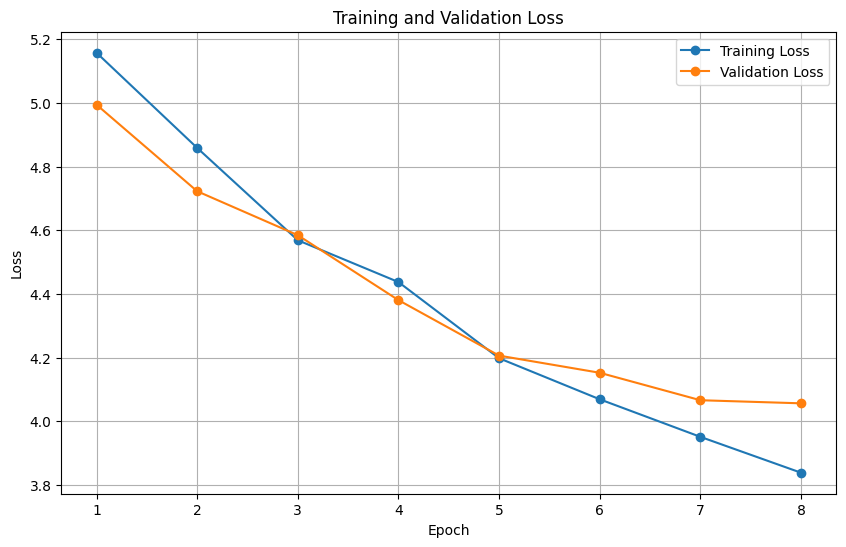}
  \caption{Training vs Validation Loss of ViT B16-BART Model}\label{fig8}
\end{figure}

\begin{figure}[h]
  \centering
  \includegraphics[width=0.75\textwidth]{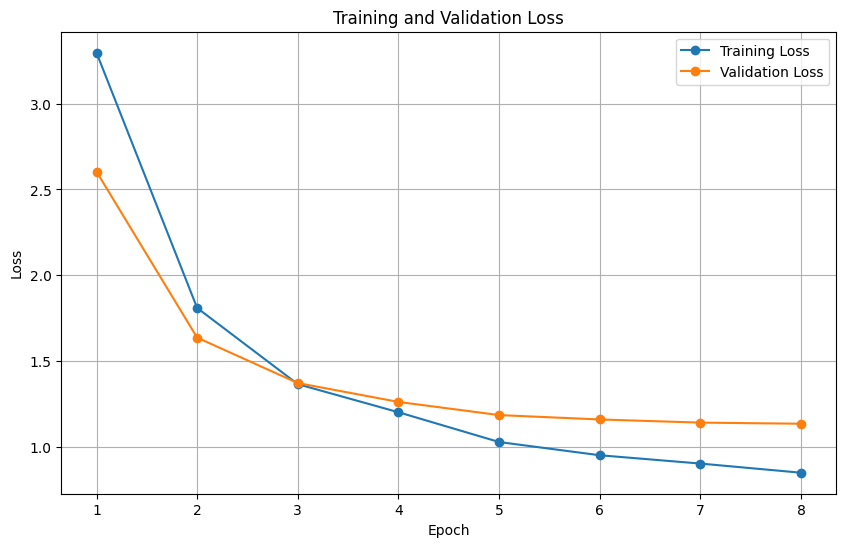}
  \caption{Training vs Validation Loss of SWIN-BART Model}\label{fig9}
\end{figure}

\begin{figure}[h]
  \centering
  \includegraphics[width=0.75\textwidth]{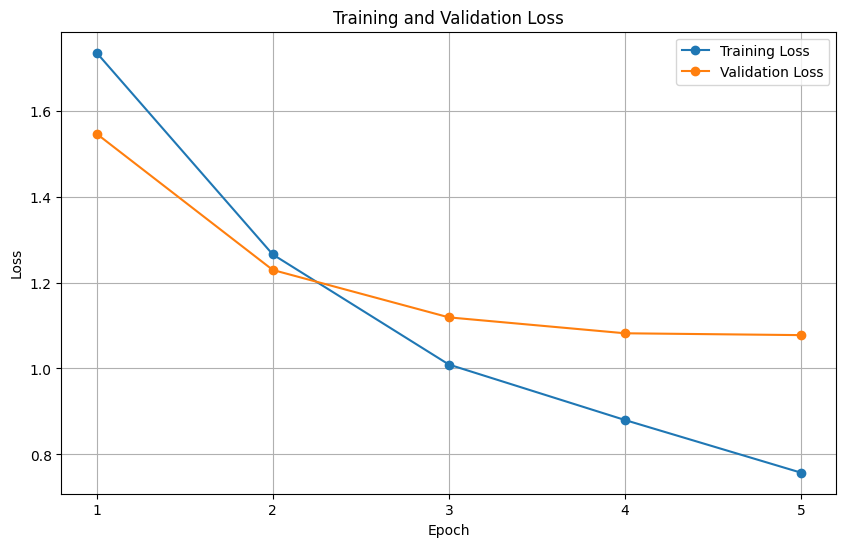}
  \caption{Training vs Validation Loss of SWIN-GPT-2 Model}\label{fig10}
\end{figure}

The comparative analysis establishes SWIN-BART as the best-performing model among the four. It consistently achieved superior results across all evaluated metrics.

Figures \ref{fig7}, \ref{fig8}, \ref{fig9} and \ref{fig10} illustrate the training and validation loss over different epochs for the ViT-B16-GPT-2, ViT-B16-BART, SWIN-BART and SWIN-GPT-2 models respectively. We observe a consistent drop in both training and validation losses for all the models.

\begin{figure}[h]
  \centering
  \includegraphics[width=0.85\textwidth]{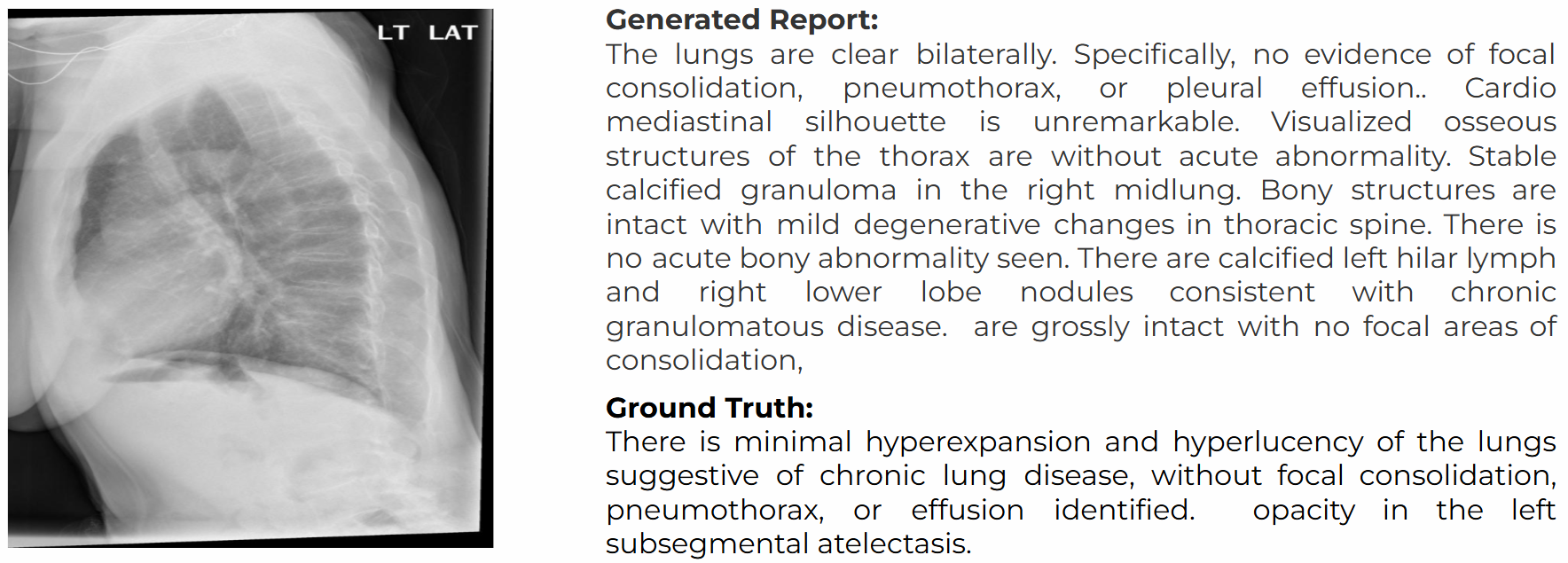}
  \caption{Generated report sample from our implementation with attached ground truth}\label{fig11}
\end{figure}

Fig. \ref{fig11} shows an example report generated through our implementation. From the attached ground truth, it is clearly visible that our model is capable of generating accurate and coherent report from a given X-ray image.

\section{Conclusion and Future Work}\label{sec6} 
In this paper we have evaluated different combinations of Vision Transformer and SWIN Transformer along with BART and GPT-2 for multimodal tasks of X-ray report generation. Our experiment finds the best performance from the SWIN-BART model in the task of accurate and coherent report generation from input Chest X-ray images. Due to computational limitations, we had to work with fixed length report generation. This is one of the limitations of our work expected to be overcome in future research. There is a lack of a more robust dataset for this generation task. This should also be addressed in the future work to get more reliable and accurate models for report generation.

\backmatter

\bibliography{sn-bibliography}

\end{document}